\documentclass[conference]{IEEEtran}
\IEEEoverridecommandlockouts
\usepackage{cite}
\usepackage{amsmath,amssymb,amsfonts}
\usepackage{algorithmic}
\usepackage{graphicx}
\usepackage{textcomp}
\usepackage{xcolor}
\usepackage{tabularx}
\usepackage{makecell}
\usepackage{color}
\usepackage{stfloats}
\usepackage{algorithmic}
\usepackage{booktabs}

\def\BibTeX{{\rm B\kern-.05em{\sc i\kern-.025em b}\kern-.08em
    T\kern-.1667em\lower.7ex\hbox{E}\kern-.125emX}}
\begin{document}

\title{Official-NV: An LLM-Generated News Video Dataset for Multimodal Fake News Detection}

\author{\IEEEauthorblockN{1\textsuperscript{st}Yihao Wang}
\IEEEauthorblockA{\textit{School of Computer Science and Technology} \\
\textit{Soochow University}\\
Suzhou, China \\
20235227104@stu.suda.edu.cn}\\
\IEEEauthorblockN{3\textsuperscript{rd}Zhong Qian$^{\ast}$}
\IEEEauthorblockA{\textit{School of Computer Science and Technology} \\
\textit{Soochow University}\\
Suzhou, China \\
qianzhong@suda.edu.cn}
\and
\IEEEauthorblockN{2\textsuperscript{nd}Lizhi Chen}
\IEEEauthorblockA{\textit{School of Computer Science and Technology} \\
\textit{Soochow University}\\
Suzhou, China \\
20234027010@stu.suda.edu.cn}\\

\IEEEauthorblockN{4\textsuperscript{th}Peifeng Li}
\IEEEauthorblockA{\textit{School of Computer Science and Technology} \\
\textit{Soochow University}\\
Suzhou, China \\
pfli@suda.edu.cn}
\thanks{$^{\ast}$ Corresponding auther}
}

\maketitle

\begin{abstract}
News media, especially video news media, have penetrated into every aspect of daily life, which also brings the risk of fake news. Therefore, multimodal fake news detection has recently garnered increased attention. However, the existing datasets are comprised of user-uploaded videos and contain an excess amounts of superfluous data, which introduces noise into the model training process. To address this issue, we construct a dataset named Official-NV, comprising officially published news videos. The crawl officially published videos are augmented through the use of LLMs-based generation and manual verification, thereby expanding the dataset. We also propose a new baseline model called OFNVD, which captures key information from multimodal features through a GLU attention mechanism and performs feature enhancement and modal aggregation via a cross-modal Transformer. Benchmarking the dataset and baselines demonstrates the effectiveness of our model in multimodal news detection.
\end{abstract}


\section{INTRODUCTION}

With the video-oriented trend of news media and the rise of video platforms, multimodal fake news has become increasingly prevalent. This type of fake news mixes text, images, and videos to mislead the public~\cite{DBLP:journals/snam/AimeurAB23, DBLP:journals/aiopen/HuWZW22,DBLP:conf/aaai/0005BC0SXWC23,DBLP:journals/corr/abs-2311-17953,DBLP:conf/mm/JinCGZL17}. Detecting fake news videos manually is labor-intensive and inefficient, which is insufficient for the large number of emerging video platforms. Therefore, Multimodal Fake News Detection (MFND)~\cite{DBLP:journals/aiopen/HuWZW22,DBLP:conf/www/LiHBW24} is a significant subtask of Natural Language Processing (NLP) and Computer Vision (CV). By analyzing multiple modalities, we can identify and flag fabricated news more accurately, enhancing the reliability of online content.

\begin{figure}[t]
\centering
\includegraphics[width=8.5cm]{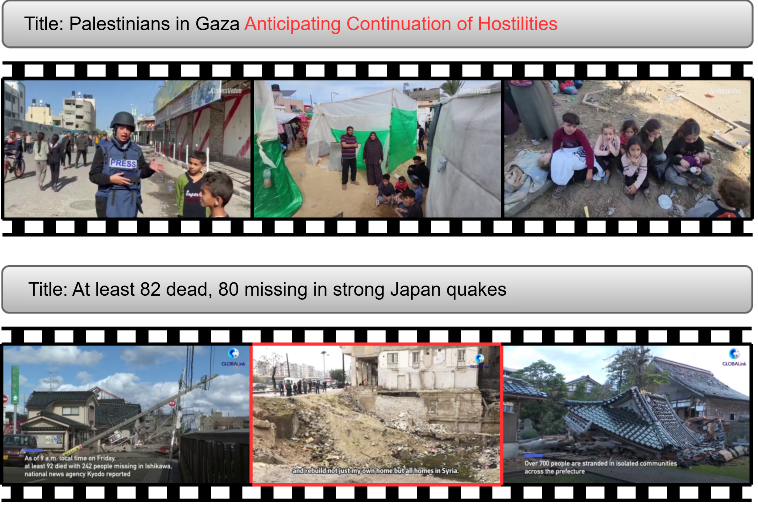}
\caption{Two examples of fake news videos. On the top the video content shows the crowds in Gaza are eager for peace, but the title is ``Anticipating Continuation of Hostilities''. On the bottom the title is ``Japan quake'', but the video is interspersed with content from the Syria quake.}
\label{fig1}
\end{figure}

In view of the importance of multimodal fake news detection, numerous datasets have been constructed. Some existing MFND datasets include FVC-2018\cite{DBLP:journals/oir/PapadopoulouZPK19}, FakeNewsNet\cite{DBLP:journals/bigdata/ShuMWLL20}, VAVD\cite{DBLP:conf/ecir/PalodPBBG19}, Liar\cite{DBLP:conf/acl/Wang17}, and MCFEND\cite{DBLP:conf/www/LiHBW24}. These datasets contain several modalities of information, such as images, text, metadata, and social context. In recent years, the rise of video social media has shifted the focus of MFND to the video modality, so researchers have proposed some MFND datasets for videos. Liu et al.\cite{DBLP:conf/eacl/LiuYS23} presented a dataset called COVID-VTS, which contains 10k English videos posted by verified users from Twitter, and proposed a fine-grained fact-checking system. Qi et al.\cite{DBLP:conf/aaai/0005BC0SXWC23} collected a total of 5538 Chinese short videos posted by users under 738 events from two short video platforms, Douyin and Kuaishou, and established a video dataset named FakeSV. It consists of 1827 real videos, 1827 fake videos, and 1884 debunked videos. Each record includes videos, titles, metadata, comments, and user information.

However, these fake news video datasets consist of unofficial, user-uploaded videos that often contain numerous duplicates or video memes. Our objective is to train on a dataset with reduced noise and enhanced quality, thereby enabling the model to achieve superior performance. To tackle this challenge, we introduce Official News Videos (Official-NV for short), a dataset comprising officially published English news videos. Specifically, Official-NV encompasses a total of 10,000 videos, evenly distributed between 5,000 authentic and 5,000 fabricated videos. Each video contains two modalities: the title and video frames. When both the title and video frames are accurate and consistent in content, the video is considered as true. When the video frames are maliciously tampered with, or the title is altered, resulting in a mismatch between the title and video frames, the video is deemed as fake. Fig. \ref{fig1} illustrates two examples of fake news videos.

To verify the superiority of our proposed dataset, we also propose \textbf{O}fficial \textbf{F}ake \textbf{N}ews \textbf{V}ideo \textbf{D}etection (\textbf{OFNVD}) as the baseline for Official-NV. This model captures important features of title and video frames through a GLU attention mechanism and then aggregates the two modalities via a cross-modal attention layer. We conducted numerical experiments using both existing models and OFNVD on the Official-NV dataset, and the results demonstrate the effectiveness of OFNVD and the superiority of Official-NV.

The main contributions are summarized as follows:
\begin{itemize}
\item[$\bullet$] We present Official-NV, a dataset consisting of officially published English news videos. It has less noise and higher video quality compared to other datasets.
\item[$\bullet$] We propose a new multi-modal baseline model, OFNVD, which captures key information from text and images through a gated attention mechanism and aggregates features through a cross-modal attention layer.
\item[$\bullet$] We conduct grouped experiments and ablation studies on Official-NV. The experimental results demonstrate that several baseline models have good performance on the proposed dataset.
\end{itemize}

\begin{table*}[t]
  \centering 
  \caption{Examples of texts that are before and after modification} 
  \label{table:1} 
  \begin{tabular}{p{0.45\textwidth}p{0.45\textwidth}} 
    \toprule 
    \textbf{Original Text} & \textbf{Modificated Text} \\ 
    \midrule 
    China's booming tea industry imbued with new momentum & China's tea industry surges forward with rejuvenated vitality \textbf{\color{blue}(TT)}\\ 
    \midrule 
    The stunning many-coloured landscapes of Xinjiang & The stunning many-coloured landscapes of Anhui \textbf{\color{red}(FT in position)}\\
    \midrule 
    Palestinian death toll from Israeli attacks in Gaza, West Bank nears 20,000 & Palestinian death toll from Israeli attacks in Gaza, West Bank more than 30,000 \textbf{\color{red}(FT in quantity)}\\
    \midrule 
    China seeks to build world's largest national park system & China aims to dismantle extensive national park network \textbf{\color{red}(FT in action)}\\
    \midrule 
    Ready... set... GO! This cat sure knows how to win a sprint race & Ready... set... GO! This dog sure knows how to win a sprint race \textbf{\color{red}(FT in object)}\\
    \bottomrule 
  \end{tabular}
\end{table*}

\section{THE OFFICIAL-NV DATASET}

In this section, we provide a detailed introduction to the construction process of Official-NV. Furthermore, we conduct data analysis and compare the differences between Official-NV and other MFND datasets.

\subsection{Data Collection}

Videos in most other MFND datasets are sourced from user-generated content platforms, such as TikTok, Twitter, and YouTube. Consequently, they often include numerous duplicate remakes or memes lacking analytical value, introducing noise into our model training. To solve this problem, we selected several official news video websites, including ChinaNews, CCTV and Xinhua to obtain news videos. Specifically, we used web crawler technology to collect news videos from the three sites from 2021 to 2024, and to minimize the noise of short, meaningless videos, we manually screened and removed all videos shorter than 15 seconds. Following this, we compute the cosine similarity between every pair of videos in the same category~\cite{DBLP:conf/emnlp/XuG0OAMZF21}, eliminating those with a similarity score exceeding 0.8, a move that can reduce the data set of video duplication. This process result in a curated collection of 2,500 news videos, classified as ``True Origin'' (TO). Each video in this category contains two modalities: title and video frames, and the information presented across these two modalities remains consistent.

\begin{table}
\centering 
\caption{Example of prompt and LLM responses} 
\label{table:2} 
\begin{tabular}{m{0.08\textwidth}p{0.32\textwidth}} 
\toprule 
\textbf{Type} & \textbf{Content}\\ 
\midrule 
Prompt & Title: ``Turkiye to introduce anti-dumping measures on solar panel imports from 5 countries'', choose a suitable Angle from the position, quantity, action and object. Change the title to another expression with the opposite meaning.\\ 
\midrule 
ChatGPT4o & Title: ``Turkiye Lifts Trade Restrictions on Solar Panel Imports from 5 Countries''
This new title presents the opposite meaning by focusing on lifting restrictions rather than imposing anti-dumping measures.\\
\bottomrule 
\end{tabular}
\end{table}

\subsection{Data Augmentation}

Inspired by methods for generating inconsistent texts and LLM image editing to extend datasets\cite{DBLP:conf/eacl/LiuYS23,DBLP:journals/corr/abs-2311-17953}, we leverage LLMs to generate fake news data and expand upon real news data after collecting 2,500 news videos. We perform data augmentation for text and video frames using two  approaches:

\textbf{Title}. Fake news often involves malicious tampering of titles, presenting information that is inconsistent with the video content. After comparing the generation effects of ERNIE, Qwen, Llama, GPT, etc., we choose to use ChatGPT4o to generate new data. Specifically, we provide prompts to ChatGPT, asking it to alter the given text to convey the same or a completely different meaning by changing aspects such as position, quantity, action and object. An example of such a conversation is presented in TABLE \ref{table:2}. Following this, we manually remove any redundant information from the responses and screened out poorly generated instances. If the modified title retains the original meaning, it is classified as a ``True Title'' (TT). If it is modified to convey the opposite meaning, it becomes a fake news video due to the inconsistency with the information conveyed by the video frames, and is therefore classified as  ``Fake Title'' (FT). TABLE \ref{table:1} displays some examples of the data after LLM generation and manual adjustment.

\begin{table*}[t]
  \centering 
  \caption{Comparison of datasets for MFND} 
  \label{table:3} 
  \begin{tabular}{ccccccc} 
    \toprule 
    {\bf Name} & {\bf Language} & {\bf Source} & \makecell{\bf Instances\\(Fake/Real)} & \makecell{\bf Title\\(Avg.length)} & \makecell{\bf Video Duration\\(Avg.length)} & {\bf Publisher} \\ 
    \midrule 
    COVID-VTS & English & Twitter & 5000/5000 & 19.2 & 26.5s & Verified Users \\ 
    FakeSV & Chinese & Douyin,Kuaishou & 1827/1827 & 32.7 & 38.5s & Users \\
        \textbf{Official-NV} & \textbf{English} & \textbf{ChinaNews, CCTV, Xinhua} & \textbf{5000/5000} & \textbf{11.3} & \textbf{69.2s} & \textbf{Official Media} \\
    \bottomrule 
  \end{tabular}
\end{table*}

\begin{figure}[t]
\centering
\includegraphics[width=1.0\linewidth]{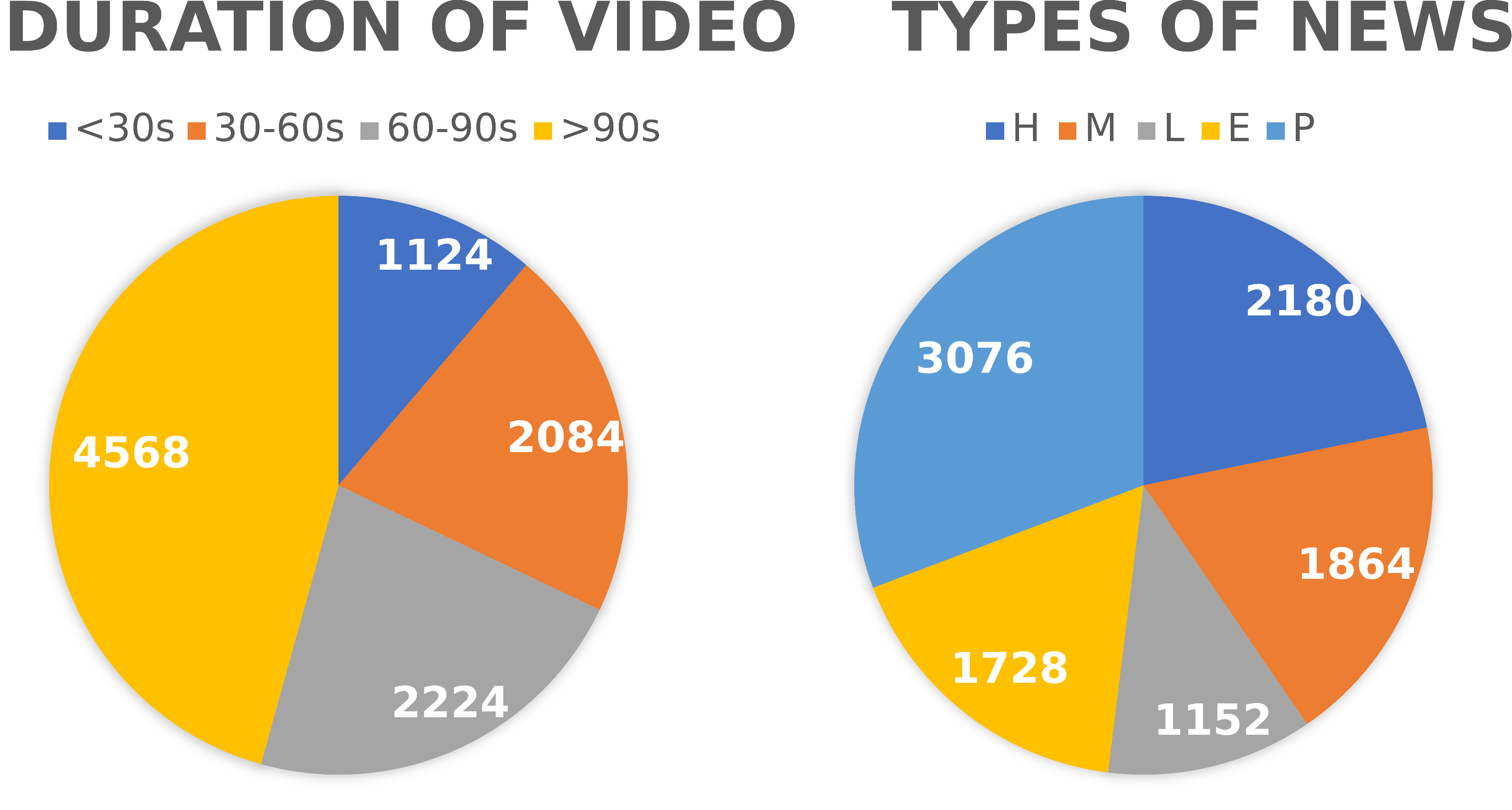}
\caption{Distributions of news videos}
\label{fig2}
\end{figure}

\textbf{Video Frame}. 
Many fake news videos splice and replace video content, using footage from another event to confuse the facts. Therefore, we decided to generate fake news videos by replacing some of the video frames. We adopt the same method used in COVID-VTS to create inconsistent videos. We compute the cosine similarity between videos to identify the one most similar to the target video and then replace part of the target video's frames with frames from this similar video, while keeping its original title intact. Due to the malicious tampering of video frames, such videos suffer from issues of inconsistency and logical coherence. These videos are labeled as ``Fake Frame'' (FF).

Following data augmentation for each video, we end up with a final dataset comprising four categories of 10,000 videos.

\subsection{Data Analysis}

The Official-NV dataset comprises 10,000 videos, derived from 2,500 official news videos through LLM data augmentation. The dataset maintains an equal number of positive and negative samples, with 5,000 each. Depending on the data generation method, the content can be segmented into four distinct categories: 1) TO: ture news videos obtained from news websites; 2) TT: true news video with title modified to a same meaning; 3) FT: fake news videos with title modified to a opposite meaning; 4) FF: fake news videos with video frames maliciously tampered. We analyze the durations of videos and categorized them into four groups based on their lengths. Additionally, we utilized LLM to classify the videos’ titles into five categories: health, military, lifestyle, education, and politics. The distribution of video durations and the classification of news types are illustrated in Fig.~\ref{fig2}.

When compared to other MFND video datasets, Official-NV offers notable advantages. In our LLM text generation process for data augmentation, we utilize a word replacement strategy, swapping words with similar or contrasting meanings. This ensures that the length of both positive and negative texts remains comparable, thereby preventing significant length disparities. With an average video duration of 69.2 seconds, Official-NV boasts considerably longer videos than other datasets, and nearly half of the videos are long videos longer than 90 seconds, allowing for a richer information content per video. Unlike datasets sourced from user-generated platforms, Official-NV is curated from officially published content, yielding videos of superior quality and reduced noise. A comparative analysis of MFND datasets is presented in TABLE~\ref{table:3}.

\section{METHOD}

\begin{figure*}[t]
\centering
\includegraphics[width=16cm]{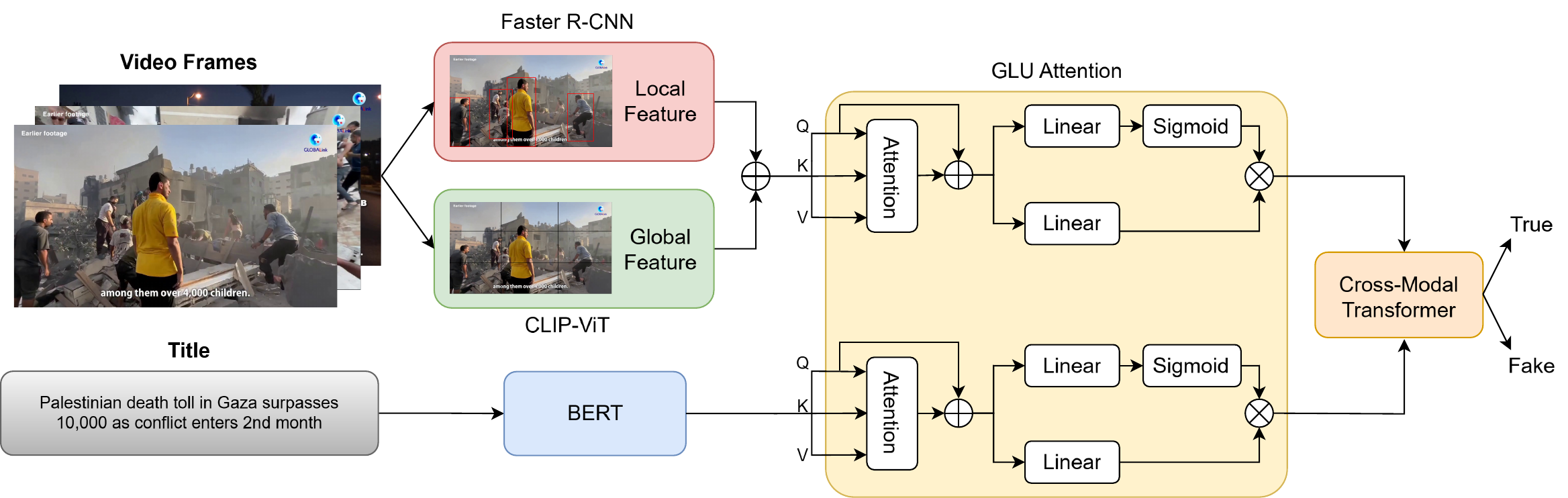}
\caption{Overview of proposed OFNVD}
\label{OFNVD}
\end{figure*}

Fig.~\ref{OFNVD} illustrates the overall architecture of the proposed OFNVD. Our baseline model can utilize the GLU Attention module to capture key information from video frame and title features, and achieve modal fusion through a cross-modal Transformer to evaluate the classification of news videos.

\subsection{Feature Extraction}

We leverage Faster R-CNN and CLIP (ViT-B/32) to encode video frames. Specifically, Faster R-CNN passes the input image through a CNN to obtain a feature map. This map encodes semantic and spatial information. The region proposal network then generates region proposals on this map, highlighting potential object locations. For each proposal, the region of interest pooling extracts a fixed-size feature vector, encoding local info for object detection and classification and forms local feature $F_{l}=[o_{1},…,o_{n}]$, where $o_{i}$ represents the feature of the $i$-th object. CLIP divides the input image into nine fixed-size patches, and each patch is flattened and mapped into a 512-dimensional vector space through linear projection. After that we obtain the video global features $F_{g}=[p_{1},…,p_{m}]$, where $p_{j}$ represents the feature of the $j$-th patch. We concatenate the two sets of features and get the final feature $F_{f}=[o_{1},…,o_{n},p_{1},…,p_{m}]$.

For the title part, we use BERT for feature extraction. We input the title text of the video into the BERT encoder to obtain the text features $F_{t}=[w_{1},…,w_{l}]$, where $w_{k}$ represents the feature of the $k$-th word and $l$ is the length of the title.

\subsection{GLU Attention Module}

Inspired by the success of attention on attention in the field of Image Caption\cite{DBLP:conf/iccv/HuangWCW19}, we adopt a similar Gated Linear Unit (GLU) attention mechanism. We use GLU to further refine and select the results of the attention layer, enabling the model to capture deeper key information from the input and thereby improving its performance in the task.

The GLU Attention module inputs the video frame features $F_{f}$ and title features $F_{t}$ into self-attention separately:
\begin{equation}
x_{f}=SelfAttention(F_{f})
\end{equation}
\begin{equation}
x_{t}=SelfAttention(F_{t})
\end{equation}
And then concatenate the output $x$ with the query:
\begin{equation}
\hat{x_{f}}=concat(x_{f},F_{f})
\end{equation}
\begin{equation}
\hat{x_{t}}=concat(x_{t},F_{t})
\end{equation}
The concatenated tensor will be sent to a gated linear unit. Specifically, it pass through two linear layers respectively. Apply Sigmoid to one of the linear layers outputs, and then calculate the element-wise product:
\begin{equation}
y_{f}=\sigma(W^{i}\hat{x_{f}}+b^{i})\otimes (W^{j}\hat{x_{f}}+b^{j})
\end{equation}
\begin{equation}
y_{t}=\sigma(W^{i}\hat{x_{t}}+b^{i})\otimes (W^{j}\hat{x_{t}}+b^{j})
\end{equation}

\subsection{Feature Aggregation}

To better integrate the features of video frames and caption texts, we employ the cross-modal Transformer adopted by Qi et al.\cite{DBLP:conf/aaai/0005BC0SXWC23} to mutually enhance the information of the two modalities. Specifically, we use a two-stream co-attention transformer to simultaneously process multimodal information, where queries from each modality are passed to the multi-head attention block of the other modality. In this way, we obtain the enhanced video frame feature $\hat{y_{f}}$ and title feature $\hat{y_{t}}$. By concatenating these two features and inputting them into the Transformer, we obtain the final aggregated feature $y_{a}$:
\begin{equation}
y_{a}=Transformer(concat(\hat{y_{f}},\hat{y_{t}}))
\end{equation}
Finally, we input the aggregated feature into a two-dimensional linear classifier to obtain the final prediction result $Output$:
\begin{equation}
Output=Classifier(y_{a})
\end{equation}

\section{EXPERIMENTS}

In this section, we conduct experiments to evaluate the performance of OFNVD and other fake news detection methods on our newly proposed Official-NV dataset. Specifically, we aim to answer the following Evaluation Questions (EQs):

{\bf EQ1}: \textit{Can the existing method demonstrate its effectiveness on the Official-NV dataset and maintain its performance across different data augmentation methods?}

{\bf EQ2}: \textit{Can the model accurately identify fake news videos when dealing with information from a single modality of data?}

{\bf EQ3}: \textit{Can the model handle the imbalance between positive and negative samples in the Official-NV dataset?}

{\bf EQ4}: \textit{What role does GLU Attention play in the process of detecting fake news in models?}

\subsection{Experimental Settings}

To address the aforementioned evaluation challenges, we devise the following experimental settings:

To improve training efficiency, we extract key features from the title of each video in the dataset using BERT\cite{DBLP:journals/corr/abs-1810-04805} before the experiment. Additionally, we employ Faster R-CNN\cite{DBLP:journals/pami/RenHG017} and CLIP\cite{DBLP:conf/icml/RadfordKHRGASAM21} to encode the visual features present in the video frames, thereby enabling a comprehensive representation of the visual content.

During the experimental phase, we select AdamW\cite{DBLP:conf/iclr/LoshchilovH19} as our optimizer, cross-entropy as our loss function, and accuracy, F1 score, precision, and recall as our evaluation metrics. This approach allow us to obtain a thorough understanding of the dataset's performance. Furthermore, we implement 5-fold cross-validation to enhance the reliability and reproducibility of our findings.


\begin{table}
  \centering 
  \caption{Results of grouped experiments} 
  \label{table:4} 
  \begin{tabular}{cccccc} 
    \toprule 
    \textbf{Methods} & \textbf{Test Data} & \textbf{Accuracy} & \textbf{F1} & \textbf{Precision} & \textbf{Recall}\\ 
    \midrule 
    SV-FEND & \makecell{Group T\\Group F\\All} & 
    \makecell{84.99\\63.40\\70.80} &
    \makecell{84.91\\54.57\\70.79} &
    \makecell{84.95\\63.40\\70.84} &
    \makecell{84.88\\57.17\\70.80}\\
    \midrule 
    TwtrDetective & \makecell{Group T\\Group F\\All} & \makecell{87.25\\63.95\\71.40} &
    \makecell{87.19\\58.42\\71.32} &
    \makecell{87.18\\62.60\\71.65} &
    \makecell{87.19\\59.21\\71.40}\\
    \midrule 
    OFNVD & \makecell{Group T\\Group F\\All} &
    \makecell{86.28\\64.22\\71.60} & 
    \makecell{86.22\\61.99\\71.51} &
    \makecell{86.20\\62.56\\71.87} &
    \makecell{86.24\\61.85\\71.60}\\ 
    \bottomrule 
  \end{tabular}
\end{table}

\subsection{Baselines}

To establish a robust and comprehensive benchmark, we conduct experiments using two representative baseline models:
\textbf{SV-FEND}\cite{DBLP:conf/aaai/0005BC0SXWC23} employs cross-modal correlation to select the most informative features and incorporates social context information for detecting false news.
\textbf{TwtrDetective}\cite{DBLP:conf/eacl/LiuYS23} integrates cross-media consistency checking to identify token-level malicious tampering across different modalities and provides explanations for its detections.


\subsection{Grouped Evaluation (EQ1)}

To demonstrate the effectiveness of the baselines on the Official-NV dataset under different data augmentation methods, we conduct grouped experiments using two baseline models and OFNVD. Specifically, we divide the dataset into three groups: Group T, consisting of 5,000 videos with modified titles (i.e., all videos labeled as TT and FT), Group F, including 5000 videos with modified video frames and all real videos crawled from news websites (i.e., all videos labeled as FF and TO), and the entire dataset containing all 10,000 videos. We perform experiments using three models on these three groups of data as well as on the entire dataset, evaluating the performance of the models across different modification methods and the overall dataset. Our experimental results are presented in Table \ref{table:4}, from which we can draw the following conclusions:
1) The accuracy of the baseline model on the Official-NV dataset is all above 0.70, and OFNVD performs best among the three baseline models, reaching 0.716, which shows the generalization of Official-NV and the superiority of the OFNVD.
2) The performance of Group T is significantly better than that of Group F, but the accuracy of Group F can still maintain above 0.62. This result indicates that text data augmentation can better improve the performance of the model, and the model can still maintain its performance when dealing with different data augmentation methods.

\subsection{Single Modal Analysis (EQ2)}

To address EQ2, we conduct a single modal analysis. Specifically, we remove the text and image modalities from the dataset separately. Since several baselines utilize cross-modal attention for multi-modal information fusion, the cross-modal attention layers used for feature enhancement are removed when a modality was absent. The results of these experiments are presented in TABLE \ref{table:5}.
From the results of the single modal analysis, we can know:
1) When the Title modality is removed from the dataset, the performance of the model drops significantly, indicating that the Title modality has a considerable impact on the model’s performance. This may be because titles often contain important summaries and information about video content, which aids the model in more accurately identifying fake news videos.
2) When the Video Frames modality is eliminated from the dataset, the model’s performance also declines slightly, suggesting that video frames provide visual information that may help the model capture details that cannot be obtained solely through text modalities.

\begin{table}
  \centering 
  \caption{Results of single modal experiments} 
  \label{table:5} 
  \begin{tabular}{ccccc} 
    \toprule 
    \textbf{Methods} & \textbf{Accuracy} & \textbf{F1} & \textbf{Precision} & \textbf{Recall}\\ 
    \midrule 
    {OFNVD} & \textbf{71.60} & \textbf{71.51} & \textbf{71.87} & \textbf{71.60} \\
    \makecell{w/o Title\\w/o Frames} &
    \makecell{60.10\\68.60} & 
    \makecell{57.97\\67.60} &
    \makecell{62.66\\71.23} &
    \makecell{60.10\\68.60}\\
    \bottomrule 
  \end{tabular}
\end{table}

\begin{table}
  \centering 
  \caption{Results of imbalanced data experiments} 
  \label{table:6} 
  \begin{tabular}{cccccc} 
    \toprule 
    \textbf{Majority} & \textbf{Minority} & \textbf{Accuracy} & \textbf{F1} & \textbf{Precision} & \textbf{Recall}\\ 
    \midrule 
    TO,TT &
    \makecell{FT\\FF} &
    \makecell{88.01\\76.39} & 
    \makecell{87.15\\75.51} &
    \makecell{87.72\\75.29} &
    \makecell{86.72\\77.35}\\
    \midrule 
    FT,FF &
    \makecell{TO\\TT} &
    \makecell{76.75\\89.34} & 
    \makecell{67.34\\88.46} &
    \makecell{68.22\\89.02} &
    \makecell{66.69\\88.01}\\
    \bottomrule 
  \end{tabular}
\end{table}

\subsection{Experiments on Imbalanced Data (EQ3)}

The number of positive and negative samples in our proposed Official-NV dataset is equal, but in reality, there may be a large deviation between the number of fake news videos and real news videos, so we conduct experiments on unbalanced data. Specifically, we remove FT or FF negative samples while keeping all positive samples, and remove TO or TT positive samples while keeping all negative samples. The experimental results are shown in TABLE \ref{table:6}. From the experimental results, we can see that when dealing with unbalanced data in Official-NV, OFNVD performs better than it does on balanced data. This is due to the fact that the model is forced to identify effective features among a few samples, thereby improving its performance.

\begin{figure}[t]
\centering
\includegraphics[width=8.5cm]{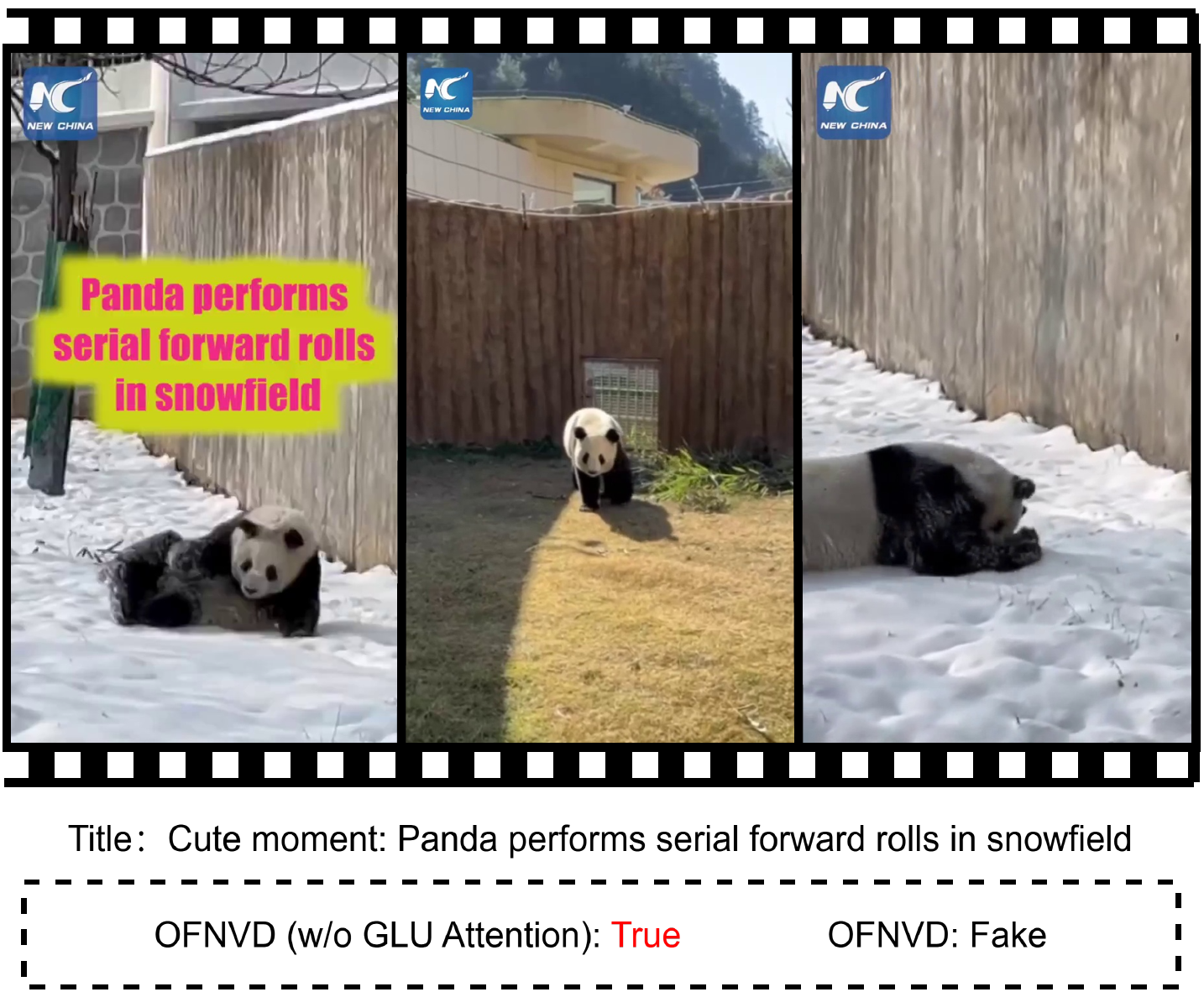}
\caption{A case in Official-NV demonstrates the role of GLU Attention}
\label{case}
\end{figure}

\subsection{Case Study (EQ4)}

Fig. \ref{case} demonstrates the role of GLU Attention in OFNVD with an example from Official-NV. We select a video with clear images, unambiguous titles, and high discriminability. In the original video, some frames of a giant panda rolling in the snow are replaced with frames of the panda playing in the grass. Without GLU Attention, OFNVD only focus on the characteristics of the panda, ignoring the incorrect information in the text about the snowfield and the absence of snow in the grassland frames. However, OFNVD with GLU Attention re-screens key information from both text and images, capturing inconsistencies between the snowfield mentioned in the text and the actual environment beyond the main subject, the giant panda. This allows for the correct identification of fake videos.

\section{CONCLUSION}

In this paper, we propose a video dataset for MFND, constructed using LLM data augmentation. Following this, we furnish a comprehensive description of the creation process of Official-NV and delve into the specifics of the data entities’ attributes and distribution. To evaluate the effectiveness of the dataset, we introduce a new baseline model, OFNVD, and conduct numerical experiments to assess the performance of various baselines, demonstrating the superiority of Official-NV. In future, we intend to explore additional avenues that hold promise for enhancing the predictive power of this dataset, with the aspiration that it will pave the way for further advancements in MFND research.

\vspace{12pt}


\begin{thebibliography}{00}

\bibitem{DBLP:journals/snam/AimeurAB23}
Aımeur, E., Amri, S. \& Brassard, G. Fake news, disinformation and misinformation in social media: a review. {\em Soc. Netw. Anal. Min.}. \textbf{13}, 30 (2023), https://doi.org/10.1007/s13278-023-01028-5


\bibitem{DBLP:journals/oir/PapadopoulouZPK19}

O.~Papadopoulou, M.~Zampoglou, S.~Papadopoulos, and I.~Kompatsiaris, ``A corpus of debunked and verified user-generated videos,'' \emph{Online Inf. Rev.}, vol.~43, no.~1, pp. 72--88, 2019.




\bibitem{DBLP:conf/eacl/LiuYS23}

F.~Liu, Y.~Yacoob, and A.~Shrivastava, ``{COVID-VTS:} fact extraction and verification on short video platforms,'' in \emph{Proceedings of the 17th Conference of the European Chapter of the Association for Computational Linguistics, {EACL} 2023, Dubrovnik, Croatia, May 2-6, 2023}, A.~Vlachos and I.~Augenstein, Eds.\hskip 1em plus 0.5em minus 0.4em\relax Association for Computational Linguistics, 2023, pp. 178--188.


\bibitem{DBLP:conf/aaai/0005BC0SXWC23}

P.~Qi, Y.~Bu, J.~Cao, W.~Ji, R.~Shui, J.~Xiao, D.~Wang, and T.~Chua, ``Fakesv: {A} multimodal benchmark with rich social context for fake news detection on short video platforms,'' in \emph{Thirty-Seventh {AAAI} Conference on Artificial Intelligence, {AAAI} 2023, Thirty-Fifth Conference on Innovative Applications of Artificial Intelligence, {IAAI} 2023, Thirteenth Symposium on Educational Advances in Artificial Intelligence, {EAAI} 2023, Washington, DC, USA, February 7-14, 2023}, B.~Williams, Y.~Chen, and J.~Neville, Eds.\hskip 1em plus 0.5em minus 0.4em\relax {AAAI} Press, 2023, pp. 14\,444--14\,452.

\bibitem{DBLP:conf/emnlp/XuG0OAMZF21}

H.~Xu, G.~Ghosh, P.~Huang, D.~Okhonko, A.~Aghajanyan, F.~Metze, L.~Zettlemoyer, and C.~Feichtenhofer, ``Videoclip: Contrastive pre-training for zero-shot video-text understanding,'' in \emph{Proceedings of the 2021 Conference on Empirical Methods in Natural Language Processing, {EMNLP} 2021, Virtual Event / Punta Cana, Dominican Republic, 7-11 November, 2021}, M.~Moens, X.~Huang, L.~Specia, and S.~W. Yih, Eds.\hskip 1em plus 0.5em minus 0.4em\relax Association for Computational Linguistics, 2021, pp. 6787--6800.



\bibitem{DBLP:conf/iclr/LoshchilovH19}

I.~Loshchilov and F.~Hutter, ``Decoupled weight decay regularization,'' in \emph{7th International Conference on Learning Representations, {ICLR} 2019, New Orleans, LA, USA, May 6-9, 2019}.\hskip 1em plus 0.5em minus 0.4em\relax OpenReview.net, 2019.





\bibitem{DBLP:journals/aiopen/HuWZW22}

L.~Hu, S.~Wei, Z.~Zhao, and B.~Wu, ``Deep learning for fake news detection: {A} comprehensive survey,'' \emph{{AI} Open}, vol.~3, pp. 133--155, 2022.




\bibitem{DBLP:journals/pami/RenHG017}

Ren, S., He, K., Girshick, R. \& Sun, J. Faster R-CNN: Towards Real-Time Object Detection with Region Proposal Networks. {\em IEEE Trans. Pattern Anal. Mach. Intell.}. \textbf{39}, 1137-1149 (2017), https://doi.org/10.1109/TPAMI.2016.2577031


\bibitem{DBLP:conf/mm/JinCGZL17}

Z.~Jin, J.~Cao, H.~Guo, Y.~Zhang, and J.~Luo, ``Multimodal fusion with recurrent neural networks for rumor detection on microblogs,'' in \emph{Proceedings of the 2017 {ACM} on Multimedia Conference, {MM} 2017, Mountain View, CA, USA, October 23-27, 2017}, Q.~Liu, R.~Lienhart, H.~Wang, S.~K. Chen, S.~Boll, Y.~P. Chen, G.~Friedland, J.~Li, and S.~Yan, Eds.\hskip 1em plus 0.5em minus 0.4em\relax {ACM}, 2017, pp. 795--816.



\bibitem{DBLP:journals/bigdata/ShuMWLL20}

K.~Shu, D.~Mahudeswaran, S.~Wang, D.~Lee, and H.~Liu, ``Fakenewsnet: {A} data repository with news content, social context, and spatiotemporal information for studying fake news on social media,'' \emph{Big Data}, vol.~8, no.~3, pp. 171--188, 2020.


\bibitem{DBLP:conf/acl/Wang17}

W.~Y. Wang, ``"liar, liar pants on fire": {A} new benchmark dataset for fake news detection,'' in \emph{Proceedings of the 55th Annual Meeting of the Association for Computational Linguistics, {ACL} 2017, Vancouver, Canada, July 30 - August 4, Volume 2: Short Papers}, R.~Barzilay and M.~Kan, Eds.\hskip 1em plus 0.5em minus 0.4em\relax Association for Computational Linguistics, 2017, pp. 422--426.


\bibitem{DBLP:conf/www/LiHBW24}

Y.~Li, H.~He, J.~Bai, and D.~Wen, ``{MCFEND:} {A} multi-source benchmark dataset for chinese fake news detection,'' in \emph{Proceedings of the {ACM} on Web Conference 2024, {WWW} 2024, Singapore, May 13-17, 2024}, T.~Chua, C.~Ngo, R.~Kumar, H.~W. Lauw, and R.~K. Lee, Eds.\hskip 1em plus 0.5em minus 0.4em\relax {ACM}, 2024, pp. 4018--4027.


\bibitem{DBLP:journals/corr/abs-2311-17953}

Z.~Sun, H.~Fang, X.~Zhao, D.~Wang, and J.~Cao, ``Rethinking image editing detection in the era of generative {AI} revolution,'' \emph{CoRR}, vol. abs/2311.17953, 2023.

\bibitem{DBLP:conf/ecir/PalodPBBG19}Palod, P., Patwari, A., Bahety, S., Bagchi, S. \& Goyal, P. Misleading Metadata Detection on YouTube. {\em Advances In Information Retrieval - 41st European Conference On IR Research, ECIR 2019, Cologne, Germany, April 14-18, 2019, Proceedings, Part II}. \textbf{11438} pp. 140-147 (2019), https://doi.org/10.1007/978-3-030-15719-7


\bibitem{DBLP:journals/corr/abs-1810-04805}

J.~Devlin, M.~Chang, K.~Lee, and K.~Toutanova, ``{BERT:} pre-training of deep bidirectional transformers for language understanding,'' \emph{CoRR}, vol. abs/1810.04805, 2018.


\bibitem{DBLP:conf/icml/RadfordKHRGASAM21}
A.~Radford, J.~W. Kim, C.~Hallacy, A.~Ramesh, G.~Goh, S.~Agarwal, G.~Sastry, A.~Askell, P.~Mishkin, J.~Clark, G.~Krueger, and I.~Sutskever, ``Learning transferable visual models from natural language supervision,'' in \emph{Proceedings of the 38th International Conference on Machine Learning, {ICML} 2021, 18-24 July 2021, Virtual Event}, ser. Proceedings of Machine Learning Research, M.~Meila and T.~Zhang, Eds., vol. 139.\hskip 1em plus 0.5em minus 0.4em\relax {PMLR}, 2021, pp. 8748--8763.

\bibitem{DBLP:conf/iccv/HuangWCW19}
Huang, L., Wang, W., Chen, J. \& Wei, X. Attention on Attention for Image Captioning. {\em 2019 IEEE/CVF International Conference On Computer Vision, ICCV 2019, Seoul, Korea (South), October 27 - November 2, 2019}. pp. 4633-4642 (2019), https://doi.org/10.1109/ICCV.2019.00473



\end{thebibliography}
\end{document}